%% file: main.tex
\documentclass[10pt,twocolumn,letterpaper]{article}

\usepackage[pagenumbers]{cvpr}

\input{preamble}

\usepackage{amsmath}
\usepackage{amssymb}
\usepackage{enumitem}

\definecolor{cvprblue}{rgb}{0.21,0.49,0.74}
\usepackage[pagebackref,breaklinks,colorlinks,allcolors=cvprblue]{hyperref}
\def\confName{CVPR}
\def\confYear{2027}

\title{GOALPOST: Where Does the Planning Gain from Test-Time Adaptation Come From?}

\author{%
Peng Zou$^{*1}$, Shicheng Yin$^{*1}$, Ziang Ni$^{*2}$, Kaixuan Yin$^{1}$, Xiyuan Zhang$^{3}$, Yanshen Xie$^{1}$, Yang Liu$^{\dagger1,4}$\\
\vspace{0.6em}
$^{1}$Sun Yat-sen University\qquad$^{2}$Delft University of Technology\\
$^{3}$Harbin Institute of Technology, Shenzhen\qquad$^{4}$X-Era AI Lab\\
\vspace{0.6em}
{\footnotesize$^{*}$Equal contribution\qquad$^{\dagger}$Corresponding author}\\
{\small \{zoup7, yinshch5, yinkx3, xieysh33\}@mail2.sysu.edu.cn,\ \ liuy856@mail.sysu.edu.cn}\\
{\small ziangni@tudelft.nl,\ \ 2025310244@stu.hit.edu.cn}\\
{\small Code: \href{https://github.com/HCPLab-SYSU/GOALPOST}{\textcolor{black}{github.com/HCPLab-SYSU/GOALPOST}}}}

\makeatletter\@namedef{ver@eso-pic.sty}{2020/10/14}\makeatother
\usepackage{pdfpages}
\begin{document}
\maketitle

\begin{abstract}
Test-time adaptation, a gradient step on recent prediction errors after each replan, raises the success of latent world models at reaching goals by 25.4 points on PushObj. One explanation is that the adapted model better predicts how the outcome changes with the action. However, on the planner's candidates, its predictions improve mostly in a shared shift, a change common to all candidates, and barely in how they differ between candidates. Under a latent goal-reaching cost, this shift has the same effect as moving the goal. GOALPOST, our closed-loop diagnostic, adds this shift to the frozen model or removes it from the adapted one. On PushObj, adding the shift recovers 94\% of the gain and removing it leaves little. Random shifts of the same norm do not help, and negating the shift hurts. We apply GOALPOST across learning rates, success criteria, search budgets, update objectives, three world models and two adaptation codebases. Adding the shift helps in every setting where adaptation does. Removing it shows that in some settings other parts of the update also contribute. Therefore, planning gains can come mostly from the shared shift, which is equivalent to re-aiming the planner at every replan. The shared shift also tracks how planning gains change with search.
\end{abstract}

\section{Introduction}

\begin{figure*}[t]
\centering
\includegraphics[width=\textwidth]{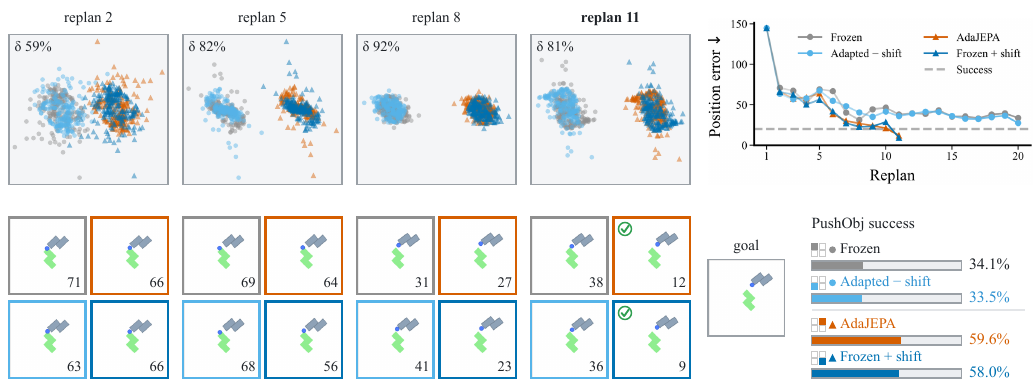}
\caption{Adding and removing the shared shift in one PushObj episode (ResNet, shape Z), run four ways from the same start. Bottom, the real states after replans 2, 5, 8 and 11 and the goal image. The number in each frame is the position error to the goal, and a check mark marks success. Only AdaJEPA and Frozen $+$ shift reach the goal. Top, the predictions for the 200 candidates under each predictor (first two principal components). Adding $\delta$ moves the frozen predictions onto the adapted ones, and removing it moves the adapted predictions back. The number in the corner is the share of the frozen-to-adapted change that $\delta$ accounts for. Top right, the position error after every replan (dashed, the success threshold). Bottom right, success in the main run (\cref{tab:main}).}\label{fig:teaser}
\end{figure*}

Latent world models often plan by search. A planner such as CEM~\cite{rubinstein2004cem} samples candidate action sequences, rolls each out in latent space, and selects the best-scoring one. Most JEPA~\cite{lecun2022path} world models score a candidate by a latent goal-reaching cost, an L1 or L2 distance between its predicted states and the embedding of a goal image~\cite{zhou2025dinowm,sobal2025pldm,assran2025vjepa2,maes2026lewm}. Planning with these scores can degrade when the test conditions differ from those seen in training~\cite{wang2026adajepa,mete2026optta}.

Test-time adaptation offers a remedy. AdaJEPA~\cite{wang2026adajepa} introduced it for JEPA world models. After each executed segment, the predictor takes a self-supervised gradient step to reduce its prediction error on recently observed transitions before replanning. On PushObj~\cite{zhou2025dinowm}, where a pusher moves an object to a goal, AdaJEPA trains its world model on four object shapes and evaluates it on seven, three of them unseen in training. We call it the ResNet model after its encoder. Its architecture follows temporal straightening~\cite{wang2026straightening}. OpTTA~\cite{mete2026optta} tests the update on the ResNet model, LeWM~\cite{maes2026lewm} and DINO-WM~\cite{zhou2025dinowm}.

Adaptation increases planning success on PushObj by 25.4 points. We reproduce the reported frozen success of all three models and the reported gains on ResNet and DINO-WM, except under salt-and-pepper noise. On LeWM we also see a gain, though smaller than reported (Sec.~4.1). The usual reading is that by reducing the prediction error on recent transitions, the update makes the adapted model better predict how the outcome changes with the action and lets the planner rank its candidates better. The planner, however, uses the model only to roll out its own candidates from the current state. An update can change these predictions in two ways. It can move all of them by the same vector, which we call the shared shift, or change how they differ between candidates, which we call the remainder. Under a latent goal-reaching cost, the differences between candidates determine which candidate scores best. The shared shift does not change the differences between the predictions but has the same effect as moving the goal, which can still change the choice (Sec.~3.3). As world models for control are judged mainly by planning success~\cite{zhou2025dinowm,sobal2025pldm,maes2026lewm,zhang2026worldinworld}, we need to know which of the two changes produces the gain. For test-time adaptation, this has not been tested.

We find that the update acts mainly as a shared shift (\cref{fig:teaser}). On the candidates the planner scores, the update reduces the prediction error mostly in its shared part and hardly in its action-dependent part, how the error differs between candidates (Sec.~4). To test whether the gain goes through the shift, we intervene in closed loop with GOALPOST, a diagnostic that adds the shared shift to the frozen model or removes it from the adapted one. On PushObj, adding the shift recovers 94\% of the gain and removing it leaves little. Random shifts of the same norm do not help, and negating the shift hurts (Sec.~5). The bulk of the gain goes through the shared shift, which is equivalent to re-aiming the planner at every replan. The remainder adds little on PushObj.

The pattern extends beyond PushObj, although in some settings the remainder also carries part of the gain (Sec.~6). For planners that score candidates by a latent goal-reaching cost, a gain from test-time adaptation can thus come mostly from a shared shift. GOALPOST shows whether it does.

Our contributions are:
\begin{itemize}[leftmargin=*,itemsep=0pt,topsep=2pt]
\item We show that AdaJEPA's planning gain on PushObj comes mostly through a shared shift, which has the same effect as moving the goal. We introduce GOALPOST, a closed-loop diagnostic that adds this shift to a frozen model or removes it from an adapted one. It applies the split that Min et al.~\cite{min2026lowerror} use for the error of a fixed model to the change made by an update, and tests by intervention in closed loop whether the gain goes through the shared shift.
\item Across 32 settings that span three world models and two codebases, adding the shift helps in all 32, and removing it takes away most of the gain in 18.
\item We show that the shared shift also tracks how planning gains change with search. With the shift, the frozen model matches or outperforms its unshifted counterpart given 3--10$\times$ as many CEM iterations. Manually amplifying the shift yields gains that diminish faster as search increases. We suggest reporting the recovered fraction at several search budgets.
\end{itemize}

\section{Related work}

\begin{figure*}[t]
\centering
\includegraphics[width=\textwidth]{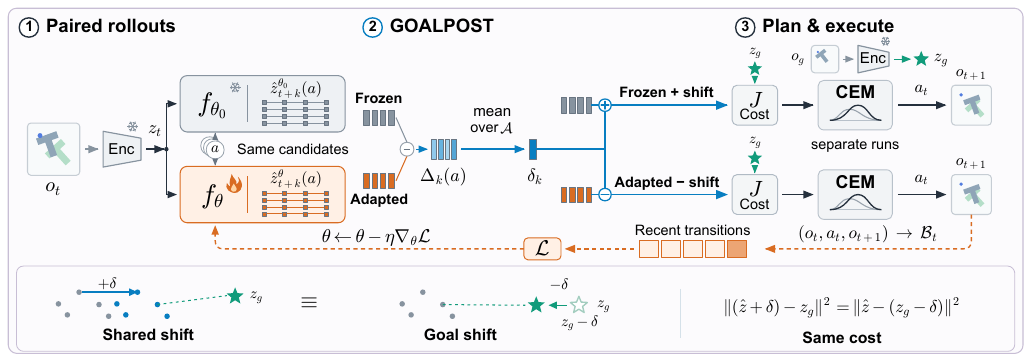}
\caption{GOALPOST in one replan. (1) The frozen and the adapted predictor roll out the same planner candidates. (2) GOALPOST computes the shared shift from their difference and runs two arms, Frozen $+$ shift and Adapted $-$ shift, each in its own closed loop. (3) CEM plans toward the goal with each arm's predictions. Orange, AdaJEPA's update on the most recent transitions. Bottom, adding the shift to every prediction has the same effect on the cost as moving the goal (Sec.~3.3).}\label{fig:method}
\end{figure*}

\paragraph{Prediction accuracy and decision quality} A more accurate model does not necessarily plan better. One-step likelihood in model-based reinforcement learning~\cite{lambert2020objective}, video-prediction metrics~\cite{tian2023vp2} and one-step error across fine-tuning checkpoints of LeWM~\cite{cheung2026onestep} can all track planning success poorly. Planning also depends on how errors differ between candidates. Dynamics error can change the ordering of candidate actions~\cite{zhang2026jepattt}. Low error on recorded transitions can coexist with poor discrimination between candidate actions~\cite{qiu2026adwm}. Dueling World Models~\cite{li2026dueling}, Counterfactual Quotient Models~\cite{chen2026cqm} and the concurrent Do-JEPA~\cite{resani2026dojepa} separate the part of a prediction common to all actions from the action-dependent part. When a candidate's score is linear in its prediction, the shared part cannot change the ranking~\cite{chen2026cqm}. Closest to us, Min et al.~\cite{min2026lowerror} note that under a squared goal distance the shared part can change the ranking through its projection onto the differences between candidates (Sec.~3.3). They split a fixed model's error on a candidate pool into a shared center and action-relative responses, and find that in six of sixteen model--task cells, repairing the responses helps action selection more than repairing the center. We instead split the change that a test-time update makes to the predictions, and test in closed loop whether its planning gain goes through the shared shift or through the remainder. The two results need not conflict, since their oracles ask which part of the error is worth repairing with realized outcomes, while we ask which part a gradient step on recent transitions changes (App.~F.1).

\paragraph{Test-time adaptation} In image classification, corrections applied identically to every input already improve robustness to image corruptions. Examples are re-estimating batch-normalization statistics on test data~\cite{schneider2020improving,nado2020evaluating} and updating the affine parameters of normalization layers as well~\cite{wang2021tent}. Adapting a dynamics model online while planning with it is much older~\cite{thrun1990adaptive}. In model-based reinforcement learning, a gradient step on the most recent transitions before each replan, akin to dynamic evaluation~\cite{krause2018dynamic}, is a baseline in GrBAL~\cite{nagabandi2019grbal} and MOLe~\cite{nagabandi2019mole}. For latent world models, AdaJEPA~\cite{wang2026adajepa} takes a self-supervised gradient step on recent transitions after every replan, JEPA-TTT~\cite{zhang2026jepattt} keeps these updates across episodes, and SimDist~\cite{levy2026simdist} fine-tunes only the latent dynamics but plans with learned reward and value heads. OpTTA~\cite{mete2026optta} adds a planning-cost term to AdaJEPA and argues that low prediction error need not translate into better planning. Residual modules around or inside a frozen predictor~\cite{soni2026sandwich,suliga2026residual} or on its action embedding~\cite{soni2026alra} make this adaptation cheaper. We take the update as given and identify which part of its change carries the gain.

\paragraph{Bias correction and search in control} Offset-free MPC augments the model with a constant disturbance, estimated from the mismatch between predicted and measured outputs, so that the closed loop reaches the set point without steady-state offset~\cite{muske2002offsetfree,pannocchia2003offsetfree}. The shared shift has the same form, but it arises from a general gradient step on prediction error and is not built into the model. As a result, the offset-free guarantees do not apply to it. Search also interacts with model error. More efficient trajectory optimizers exploit model errors more readily~\cite{boney2019regularizing}, and planners exploit model error that persists across their iterations~\cite{song2026dither}. In latent world models, further optimization can keep lowering the predicted cost while success falls~\cite{zhang2026efe}. We ask how the gain of test-time adaptation changes as CEM search gets stronger.

\section{Test-time adaptation and the shared shift}\label{sec:method}

\paragraph{3.1 Planning and adaptation}
We follow the notation of AdaJEPA~\cite{wang2026adajepa}. An encoder maps an observation $o_t$ to a latent state $z_t$. A predictor $f_\theta$ maps $z_t$ and an embedded action $u_t$ to the next latent state. Rolling the predictor out from $z_t$ under a candidate action sequence $a$ gives predictions $\hat z_{t+k}(a)$ for $k=1,\dots,H$. A planner scores each candidate by the latent goal-reaching cost
\begin{equation}J(a)=\sum_{k=1}^{H}\alpha_k\, d\big(\hat z_{t+k}(a),\,z_g\big),\end{equation}
where $d$ is a squared distance between latent states, $z_g$ is the embedding of the goal image and $\alpha_k$ are per-step weights. App.~A.1 gives $d$ and $\alpha_k$ for each model. CEM samples candidates, keeps the lowest-cost elites, and executes the first actions of the mean of the last iteration's elites before replanning.

Test-time adaptation updates the predictor after each executed segment (\cref{fig:method}, orange). With $\mathcal B_t$ the five most recent transitions,
\begin{gather}\mathcal L(\theta)=\frac{1}{|\mathcal B_t|}\sum_{(o_i,a_i,o_{i+1})\in\mathcal B_t}\big\lVert f_\theta(z_i,u_i)-\mathrm{sg}(z_{i+1})\big\rVert_2^2,\\\theta\leftarrow\theta-\eta\,\nabla_\theta\mathcal L(\theta).\end{gather} We call the predictor with the pretrained weights $\theta_0$ the \emph{frozen model} and the one with the updated weights $\theta$ the \emph{adapted model}.
Only the last predictor block and the final LayerNorm are updated (the form above is schematic). Freezing the final LayerNorm, whose bias shifts every prediction alike, leaves the gain on ResNet unchanged. Updating only that LayerNorm recovers little of it (App.~B.3). It follows that the shift we study is not simply that bias. The weights are reset to the pretrained $\theta_0$ at the start of every episode. We keep the encoder frozen, which barely changes adapted success on ResNet (64.0\% against 63.9\% over seven shapes).

\paragraph{3.2 What an update changes at the planning interface}
The planner sees the model only through its predictions for the candidates it scores (\cref{fig:method}, panel 1). At each replan and prediction step, the update's change to these predictions is
\begin{equation}\Delta_k(a)=\hat z^{\theta}_{t+k}(a)-\hat z^{\theta_0}_{t+k}(a).\end{equation}
We split $\Delta_k(a)$ over a set $\mathcal A$ of 64 reference candidates into a part shared by all candidates and a remainder that depends on the action:
\begin{equation}\delta_k=\frac{1}{|\mathcal A|}\sum_{a\in\mathcal A}\Delta_k(a)=\langle\Delta_k\rangle_{\mathcal A},\qquad \tilde\delta_k(a)=\Delta_k(a)-\delta_k.\end{equation}
We call $\delta_k$ the \emph{shared shift} and $\tilde\delta_k(a)$ the \emph{remainder}. The shift is recomputed at every replan from the first 64 of the 200 samples of the first CEM iteration. Computing it from the last iteration's candidates or elites instead gives nearly the same direction and recovers a similar share of the gain (App.~B.1).

\paragraph{3.3 A shared shift moves the goal}
As $J$ depends on a prediction only through its difference from the goal,
\begin{equation}d\big(\hat z_{t+k}(a)+\delta_k,\;z_g\big)=d\big(\hat z_{t+k}(a),\;z_g-\delta_k\big)\end{equation}
for every candidate. In other words, adding the shared shift to all predictions has the same effect as planning toward the moved goals $z_g-\delta_k$, one per step, with the unshifted predictions. The shift does not change the predicted differences between candidates. For a squared distance $d$ with fixed per-dimension weights $W$ and $\hat z=\hat z_{t+k}(a)$,
\begin{equation}d(\hat z+\delta_k,z_g)=d(\hat z,z_g)+2\delta_k^\top W(\hat z-z_g)+\lVert\delta_k\rVert_W^2.\end{equation}
The middle term differs between candidates, so the shift \emph{can} change which candidate scores best. Min et al.~\cite{min2026lowerror} derive the same cross term for the shared part of a fixed model's error, whereas here it is the shared shift that an update adds. Neither the identity nor the split of $\Delta_k(a)$ implies that the gain comes through the shift. Testing whether it does requires intervening in closed loop.

\paragraph{3.4 GOALPOST}
GOALPOST is a closed-loop diagnostic that adds the shared shift to the frozen model or removes it from the adapted one. Each arm is a separate closed-loop run that resets to $\theta_0$ at the start of every episode. At every replan, each arm computes $\delta_k$ from an adapted predictor that is updated only on the transitions that arm has executed.
\begin{itemize}[leftmargin=*,itemsep=0pt,topsep=2pt]
\item \textbf{Frozen $+$ shift} plans with the frozen predictions plus $\delta_k$. The adapted predictor is used only to compute the shift.
\item \textbf{Adapted $-$ shift} plans with the adapted predictions minus $\delta_k$.
\end{itemize}
The shift is added to or subtracted from the predictions after the rollout and is not fed back into the predictor. At each replan, the adapted predictions minus the shift equal the frozen predictions plus the remainder, $\hat z^{\theta}_{t+k}(a)-\delta_k=\hat z^{\theta_0}_{t+k}(a)+\tilde\delta_k(a)$. Adapted $-$ shift keeps only the part of the update that differs between candidates. As controls, we replace the shift with substitutes, mainly a \textbf{random shift} with the same norm at each step but a random direction, added to the frozen model (others in Sec.~5.2).
As the arms execute different actions, their states, update data and candidates diverge after the first replan. We call the success rate of the adapted model minus that of the frozen model the \emph{gain}. Adding the shift tests whether it is sufficient for the gain, and removing it tests whether the gain needs it. We report the \textbf{recovered fraction}, the gain of Frozen $+$ shift divided by the gain of the adapted model, and the \textbf{remainder gain}, the success rate of Adapted $-$ shift minus that of the frozen model. A recovered fraction above 100\% means that Frozen $+$ shift plans better than the full update, that is, the remainder hurts. Both quantities are read relative to the gain, which makes GOALPOST informative only where adaptation clearly helps. The recovered fraction is not an additive share of a causal effect. Computing the shift requires an adapted model, so Frozen $+$ shift saves no computation.

\paragraph{3.5 Measuring prediction accuracy}
We measure prediction error in three places. \emph{Next-segment error} compares, within an episode, the adapted and frozen predictions of each newly executed segment before the update uses it. \emph{Held-out error} uses 85 validation segments per shape that the update never trains on. We report the one-step error and the five-step open-loop error under the recorded actions, relative to the frozen model. \emph{Error on the planner's candidates} uses two sets of candidates (App.~C). The \emph{sample set} is the first 16 candidates of the first CEM iteration at the 5th and 10th replan. The \emph{elite set} is the union of the 16 best candidates of the frozen and the adapted model at the first CEM iteration. We replay them in the simulator from the current state and encode the outcomes as $z_{t+k}(a)$. Writing $e_k(a)=\hat z_{t+k}(a)-z_{t+k}(a)$ for the prediction error, we split it over these candidates in the same way as $\Delta_k(a)$, into a shared part $\bar e_k$, its mean over the candidates, and an action-dependent part $\tilde e_k(a)=e_k(a)-\bar e_k$. The two parts are the counterparts of the shared shift $\delta_k$ and the remainder $\tilde\delta_k(a)$. The action-dependent part of a prediction is its difference from the mean over the candidates. Here, $\tilde e_k(a)$ is the error in that part. We measure the size of each part of the error by its squared norm, summed over latent dimensions and averaged over rollout steps, so that the two add up to the total squared error.

\paragraph{3.6 Protocol and statistics}
Arms are compared episode by episode within seeds. Intervals resample tasks, segments, source trajectories or source demonstrations jointly for all arms. We say the shift carries a gain when adding it recovers most of the gain and removing it leaves little. The paired tests are Holm-corrected within each batch. With margins fixed before the data were seen (App.~A.3), we classify removal as \emph{gone} when the remainder gain lies below 30\% of the gain with 95\% confidence, and otherwise as \emph{part left} when its 95\% interval lies above zero and as \emph{imprecise} when the interval covers both zero and 30\% of the gain.
For the batches we report, the analyses were fixed before their outcomes were seen. Analyses added after seeing their outcomes are marked post hoc. The 4$\times$ learning rate and the coefficient of the residual feedback in Sec.~7.1 were chosen on development seeds.

\section{The adapted model predicts better mostly through a shared shift at the original learning rate}\label{sec:setup}

\paragraph{4.1 Setting}
Our main model is the ResNet model, AdaJEPA's released PushObj shape-shift checkpoint (SmallResNet encoder, ViT predictor), trained on the shapes T, L, Z and +, and evaluated on all seven shapes. Planning uses the released CEM. An episode succeeds if, after any replan, the position error of the pusher and block is below 20 and the block's angle error is below 20$^\circ$, as in AdaJEPA, DINO-WM and LeWM. Adaptation uses the learning rate of the original papers unless stated otherwise, $5\times10^{-4}$ on ResNet and DINO-WM and $5\times10^{-5}$ on LeWM. We also test a 4$\times$ learning rate. Our frozen success and gain are 34.1\% / +25.4 on ResNet (reported 32.0\% / +27.7), 79.1--81.4\% / +3.1--6.9 on LeWM (80.0\% / +8.8) and 74.0\% / +5.3 on DINO-WM (76.7\% / +0.7). App.~A.1 gives the full setup. We analyze each model in the environment of the paper that reported its gain (the ResNet model on AdaJEPA's tasks, and LeWM and DINO-WM on OpTTA's). On the ResNet model and DINO-WM, we run GOALPOST in the conditions where the original paper reports an adaptation gain of at least 10 points, and on DINO-WM also under darkening. Sec.~6.3 gives the LeWM conditions. App.~A.5 compares every reproduced condition with the original paper.

\begin{figure*}[t]
\begin{minipage}[t]{4.48in}\vspace{0pt}
\centering\footnotesize\setlength{\tabcolsep}{1.6pt}
\captionof{table}{Adding and removing the shared shift in the main setting of each model (App.~E.5 lists every setting, and \cref{fig:forest} shows 36 of them). Success in \%, gain and remainder gain in points, brackets give 95\% intervals. \emph{Recovered} is the gain of Frozen $+$ shift as a share of the gain, and above 100\% Frozen $+$ shift gains more than the full update. \emph{Remainder} is Adapted $-$ shift minus Frozen.}\label{tab:main}
\begin{tabular}{l c c c c c r@{\,}l r@{\,}l}
\toprule
Setting & Frozen & Adapted & Gain & \shortstack{Frozen\\$+$ shift} & \shortstack{Adapted\\$-$ shift} & \multicolumn{2}{c}{\shortstack{Recovered\\(\%)}} & \multicolumn{2}{c}{Remainder} \\
\midrule
\multicolumn{10}{l}{\emph{ResNet, PushObj}} \\
\hspace{0.6em}Main & 34.1 & 59.6 & +25.4 & 58.0 & 33.5 & 94 & {\scriptsize[87, 101]} & $-$0.6 & {\scriptsize[$-$2.4, 1.2]} \\
\hspace{0.6em}Learning rate $4\times$ & 33.9 & 63.8 & +29.9 & 62.5 & 35.4 & 96 & {\scriptsize[90, 102]} & +1.5 & {\scriptsize[$-$0.2, 3.1]} \\
\multicolumn{10}{l}{\emph{LeWM, PushT}} \\
\hspace{0.6em}Released checkpoint & 80.1 & 86.2 & +6.1 & 86.8 & 80.1 & 111 & {\scriptsize[102, 121]} & 0.0 & {\scriptsize[$-$0.6, 0.7]} \\
\multicolumn{10}{l}{\emph{DINO-WM, PushT, AdaJEPA update}} \\
\hspace{0.6em}Salt-and-pepper 0.1\% & 47.3 & 64.0 & +16.7 & 70.7 & 34.0 & 140 & {\scriptsize[103, 217]} & $-$13.3 & {\scriptsize[$-$21.3, $-$5.3]} \\
\hspace{0.6em}Gamma 0.2 & 53.3 & 64.7 & +11.3 & 68.7 & 57.3 & 135 & {\scriptsize[95, 250]} & +4.0 & {\scriptsize[$-$3.3, 11.3]} \\
\hspace{0.6em}Translation 5 px & 42.7 & 55.3 & +12.7 & 56.0 & 40.0 & 105 & {\scriptsize[52, 192]} & $-$2.7 & {\scriptsize[$-$10.0, 4.7]} \\
\hspace{0.6em}Contrast $\times 2$ & 16.0 & 26.0 & +10.0 & 32.0 & 12.0 & 160 & {\scriptsize[105, 288]} & $-$4.0 & {\scriptsize[$-$8.7, 0.0]} \\
\hspace{0.6em}Darkening $\times 0.1$ & 16.7 & 29.3 & +12.7 & 38.7 & 12.7 & 174 & {\scriptsize[111, 340]} & $-$4.0 & {\scriptsize[$-$10.7, 2.7]} \\
\bottomrule
\end{tabular}
\end{minipage}\hfill
\begin{minipage}[t]{2.10in}\vspace{0pt}
\centering\includegraphics{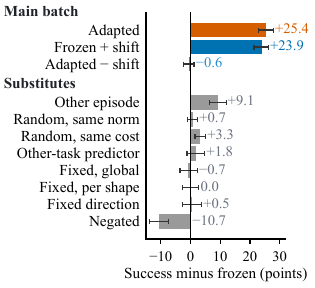}
\captionof{figure}{Only the shared shift recovers the gain. Success minus the frozen model on PushObj at the original learning rate, in points, with 95\% intervals. Top, the main batch. Bottom, substitutes for the shift.}\label{fig:subst}
\end{minipage}
\end{figure*}

\paragraph{4.2 Success and prediction error at two learning rates}
Adaptation raises success over the frozen model by 11.7 to 38.3 points on the seven shapes. At the original learning rate, the adapted model's next-segment error is lower than the frozen model's. Its held-out error changes by $-2\%$ to $+4\%$ for one step and $-1\%$ to $+6\%$ for five steps. At $4\times$, both errors are clearly worse than the frozen model's. The one-step held-out error rises by 29\% to 106\% on Z, L, square and I. The next-segment error exceeds the frozen model's on three of the four shapes. However, the gain on these four shapes is 29.9 points at $4\times$, against 25.4 at the original rate (\cref{tab:main}).

\paragraph{4.3 On the planner's candidates, predictions improve mostly through a shared shift}
The planner scores its own candidates, not these trajectories. On the planner's candidates, replayed in the simulator from the current state, adaptation lowers the prediction error by 14.2\%. Split as in Sec.~3.5, the shared part $\bar e_k$ falls by 22.3\% (post hoc) and the action-dependent part $\tilde e_k(a)$ by only 3.0\%. By construction, the frozen predictions plus the shared shift have almost the same $\bar e_k$ as the adapted predictions, so the drop in $\bar e_k$ is not separate evidence that the shift matters. What the data add is that the frozen model's $\bar e_k$ points against the shift (mean per-state cosine $-$0.49, post hoc), which means that the shift moves the predictions toward the realized outcomes. The shift still differs from the oracle shift $b^\ast$ that would cancel the frozen model's $\bar e_k$ (Sec.~7.1). On the sample set, the ranking of candidates by predicted cost improves only slightly (Spearman correlation with the realized costs up by 0.015, of which the shift alone gives 0.013), but on the elite set it rises by 0.104, of which the shift alone gives 0.103 (App.~C). At 4$\times$, the ranking is worse than the frozen model's on the same candidates (Spearman correlation lower by 0.022 [0.017, 0.028]).

\section{Why does adaptation help?}\label{sec:resnet}

\begin{figure*}[t]
\centering
\includegraphics[width=\textwidth]{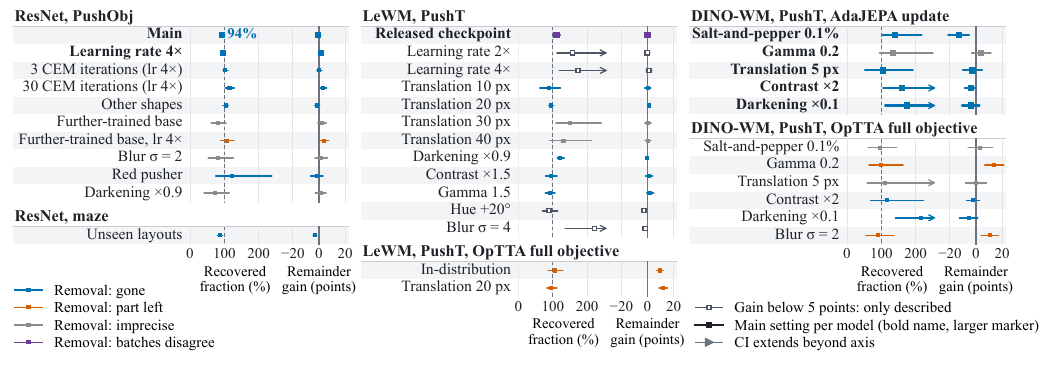}
\caption{Adding and removing the shared shift in the 40 settings of Sec.~6, except four LeWM checkpoints we trained ourselves. Left of each panel: the recovered fraction (dashed line: 100\%). Right: the remainder gain in points (solid line: 0). Squares are point estimates and bars 95\% intervals. Colors give how removal is classified (App.~A.3), and open markers show settings whose gain is below 5 points, which are only described. Bold names mark the main setting of each model (\cref{tab:main}). App.~E.5 gives the numbers.}\label{fig:forest}
\end{figure*}

\paragraph{5.1 Adding and removing the shift}
On Z, L, square and I (\cref{tab:main}), adaptation raises success by 25.4 points over the frozen model's 34.1\%. Frozen $+$ shift recovers 94\% [87\%, 101\%] of this gain. The remainder gain of Adapted $-$ shift is $-$0.6 points [$-$2.4, 1.2]. Almost all of the gain goes through the shift. The shift also carries the gain on three further shapes (104\%, \cref{fig:forest}) and at the 4$\times$ learning rate (96\%). At 4$\times$, this holds even though the prediction error on the planner's candidates does not fall (+2.0\% [0.1\%, 4.1\%]), so the shift carries the gain even when the predictions do not become more accurate. It also holds on each shape separately, although the gain itself ranges from about 18 points on the unseen shapes to 33 on the training shapes.

\paragraph{5.2 Size, shape or direction of the shift}
AdaJEPA attributes the gain to specialization to the current shape~\cite{wang2026adajepa}. We test whether the gain can also be recovered by a shift that matches its size or depends only on the shape (\cref{fig:subst}). \emph{Matching the size of the shift is not enough.} A random shift with the same norm does not detectably raise success. A shift scaled to perturb the cost differences between candidates as much as the actual shift adds only 3.3 points. \emph{A shape-level correction is not enough either.} Shifts taken from other episodes of the same shape raise success by 9.1 points, about a third of what the actual shift adds, while a fixed vector per shape, a single global vector or a fixed direction with the current length add no detectable success. Loading the predictor adapted on another task of the same shape adds no detectable success either, whereas the predictor adapted on the same task recovers 108\% of the gain. Negating the shift lowers success by 10.7 points relative to the frozen model. It is the direction of the shift, computed for the current state at every replan, that carries the gain.

\paragraph{5.3 The result is not implied by the identity}
If the update changed predictions almost only through the shared shift, ``adding works, removing fails'' would restate one fact. That would require the remainder to be negligible for planning. \emph{The remainder is not negligible.} The planner sees only cost differences between candidates. The remainder perturbs them by 0.114 of the spread of the frozen costs, against 0.145 for the shift and 0.034 for a random shift of the same norm, changing about 7\% of the elites. \emph{The remainder adds no detectable gain.} Without it, Frozen $+$ shift recovers 94\% of the gain, and with it alone, Adapted $-$ shift keeps no detectable gain (Sec.~5.1). If the remainder mattered for success, either outcome could have failed. Hence, adding the shift and removing it are two separate findings.

\paragraph{5.4 The success criterion}
Counting the block alone as success cuts the gain by more than half, but the shift still recovers 96\% of it, and 105\% at 4$\times$.

\section{Beyond PushObj, and when adaptation hurts}\label{sec:scope}
We now ask whether the result of Sec.~5 holds when the base model, the task and images, the world model and codebase, the update objective or the search budget changes, and run GOALPOST in all 40 settings where adaptation helps and removal is tested (\cref{fig:forest}, App.~E.5). In the 32 where the gain is at least 5 points, adding the shift helps in all of them, and removing it takes away most of the gain in 18. A clear part of the gain remains in 5, where the remainder also contributes. In 8 the remainder gain is too imprecise to classify. In the last, LeWM's released checkpoint, separate batches disagree. The other 8, all on LeWM, do not reach 5 points and are only described (App.~E.5).

\paragraph{6.1 A stronger base}
The released ResNet checkpoint was saved after 3 of the 5 epochs in its training configuration, which means the result could reflect a weak model. We continued its training on data from which every trajectory of the evaluation tasks was removed. The frozen success rises to 43.3\%, while the gain of adaptation falls to 15.5 points (18.4 at 4$\times$). Frozen $+$ shift recovers 81\% of the gain (107\% at 4$\times$) and beats a random shift of the same norm at both rates. The shift carries the gain less clearly than on the released checkpoint. At 4$\times$, removing it leaves 4.1 points, about a fifth of the gain. At 1$\times$, the remainder gain, +2.4 points, is too imprecise to classify. Four LeWM checkpoints that we trained from scratch with other seeds point the same way, but their gains of 3.6 to 4.7 points are below 5 and only described.

\paragraph{6.2 Another task and shifted images}
Two further tests ask whether the result is specific to pushing an object in clean images. In PointMaze~\cite{fu2020d4rl}, AdaJEPA's released multi-maze checkpoint~\cite{wang2026adajepa}, trained on 25 maze layouts, is evaluated on 5 unseen ones. Adaptation raises the success of the CEM planner by 17.0 points. Frozen $+$ shift recovers 87\% of this gain, and the remainder gain is $-$3.2 points (\cref{fig:forest}). A random shift of the same norm also helps. Its gain of 4.7 points shows that a small part of the effect does not depend on the direction. The shift still beats it by 10.2 points.

On PushObj we also test the three image shifts for which AdaJEPA reports a gain of at least 10 points, namely blur ($\sigma = 2$), a red pusher and darkening ($\times 0.9$). Adaptation gains 13.3, 10.7 and 12.7 points. Under all three, the shift beats a random shift of the same norm. Under the red pusher, Frozen $+$ shift recovers 122\% of the gain, and the remainder gain is $-$1.7 points [$-$6.5, 3.0]. Under blur and darkening, Frozen $+$ shift recovers 82\% and 74\%, and the remainder gain, +1.7 [$-$3.0, 6.4] and +1.3 [$-$2.8, 5.4] points, is too imprecise to classify. On unseen layouts and under the red pusher, the shift carries the gain, while under blur and darkening these batches cannot tell whether the remainder also contributes.

\paragraph{6.3 Other world models}
A second world model and codebase test whether the result depends on the ResNet model. On LeWM we run the public code of OpTTA~\cite{mete2026optta}. Adaptation is still AdaJEPA's update with OpTTA's settings. Our wrapper only adds the shift to or removes it from the planner's predictions, and reproduces the original runs exactly when switched off. Our frozen success agrees with OpTTA's under six of seven image corruptions and our gain under all seven. In the largest batch on the released checkpoint, Frozen $+$ shift recovers 111\% [102\%, 121\%] of the 6.1-point gain, more than the full update does, and the remainder gain is 0.0 points [$-$0.6, 0.7]. Two smaller batches agree in their point estimates, recovering 110\% and 124\% with a remainder gain of +0.6 points, but in the smaller one the remainder gain is too imprecise to classify, which makes this the one setting where batches disagree. Under seven image corruptions with a gain of at least 5 points, Frozen $+$ shift recovers 88\% or more of the gain, and removing the shift takes away most of it in all but translation by 30 and 40 pixels, where the remainder gain is too imprecise to classify (\cref{fig:forest}). Under image corruptions the encoder stays frozen while its inputs change, which may favor a shared shift. For this reason, we treat these results as supporting evidence. At 2$\times$ and 4$\times$ the learning rate, and under hue and blur, the gain is below 5 points and is only described.

DINO-WM, a third world model with a frozen DINOv2~\cite{oquab2024dinov2} encoder, is evaluated with OpTTA's settings in distribution and under seven image corruptions. We reproduce OpTTA's frozen success in all eight conditions. Our gains are consistent with OpTTA's in all but salt-and-pepper noise, where ours is smaller. We run GOALPOST under the five corruptions with a reported gain of at least 10 points and under darkening. Adaptation helps in five of these six, all but blur. Adding the shift helps in all five, by 13.3 to 23.3 points, and the remainder gain is $-$13.3 to $-$2.7 points under salt-and-pepper noise, translation, contrast and darkening. Under gamma the remainder gain, +4.0 points [$-$3.3, 11.3], is too imprecise to classify, and a random shift of the same norm also helps (+6.0 [0.7, 12.0]).

\paragraph{6.4 Another update objective}
OpTTA's full objective adds a planning-cost term to AdaJEPA's prediction loss. With it we test whether the result depends on the update objective. We run GOALPOST with the shift computed from a copy updated with this objective, on LeWM in distribution and under translation by 20 pixels, and on DINO-WM under the six corruptions of Sec.~6.3. Adaptation gains 9.1 and 15.2 points on LeWM, and on DINO-WM 12.0 to 17.3 points, within the range OpTTA reports. Adding the shift helps in all eight settings, recovering 105\% and 96\% of the gain on LeWM and 90\% to 216\% on DINO-WM. Removal behaves differently from AdaJEPA's update. On LeWM, the remainder gain is almost the whole gain, 9.4 of 9.1 points and 12.1 of 15.2. On DINO-WM, removing the shift takes away most of the gain under contrast and darkening. Under translation and salt-and-pepper noise the remainder gain is too imprecise to classify, and under blur and gamma a clear part of the gain remains, +10.7 and +14.0 points (\cref{fig:forest}). With this objective, the shift is sufficient for the gain but not always necessary. These four settings and the further-trained base at 4$\times$ are the 5 in Sec.~6 where a clear part of the gain remains. Sec.~7.1 discusses why the remainder helps here.

\begin{figure}[t]
\centering
\includegraphics[width=\linewidth]{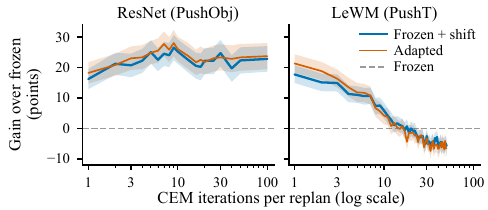}
\caption{Gain over the frozen model as CEM iterations grow (log scale), for the adapted model and for the frozen model plus the shared shift, with 95\% intervals. Every tested iteration count is shown.}\label{fig:search}
\end{figure}
\paragraph{6.5 Search budget}
We vary only the search budget, the number of CEM iterations, and keep the other planner settings (\cref{fig:search}). \emph{On PushObj, the gain persists as the search budget grows.}
From 3 to 30 iterations it changes by +0.2 points [$-3.7$, 4.1], while the frozen success rises from 23.2\% to 41.0\%. From 30 to 100 iterations, where the frozen model no longer improves, it changes by +0.3 points [$-4.5$, 5.2] (3 seeds). Adding the shift to the frozen model at 3 iterations gives 4.8 points more success than the frozen model at 30, with about a tenth of the CEM iterations. \emph{On LeWM, the gain fades as the frozen model nears its ceiling.}
In a batch of 6 seeds, the frozen success rises from 41.7\% at 1 iteration to 86.0\% at 30, while the gain falls by 23.0 points. The frozen model with the shift at 10 iterations matches the frozen model at 30 (+1.1 points [$-3.0$, 5.2]) with about a third of the CEM iterations. Frozen $+$ shift follows the adapted model across budgets on both models (\cref{fig:search}). \emph{With a larger shift, the gain fades faster.} On PushObj, scaling the shift by 1.5 and 2 makes the gain shrink from 3 to 30 iterations by 9.9 and 15.6 points more. On LeWM, halving it makes the gain shrink from 1 to 30 iterations by 10.7 points less, and doubling it by 21.1 points more.

\section{Discussion and conclusion}

\par\vspace{.5em}\noindent\textbf{7.1 What does the shift correct?}
Our results show where the gain comes from more clearly than why the shift helps. The shift may correct a systematic bias in the predictions, change how the planner scores candidates, or correct an aiming error of the frozen model and cost together. The first reading fits PushObj better than LeWM. On PushObj, adding the oracle shift $b^\ast$, computed with the simulator, does better than the shift (67.7\% against 60.0\%).
Residual feedback, which adds the mean one-step residual of recent segments to the frozen predictions with a coefficient of 1 and no parameter update, also recovers 78\% of the gain there. It is a form of the constant disturbance estimate of offset-free MPC~\cite{muske2002offsetfree,pannocchia2003offsetfree}, carried over to latent world models. In this sense, on PushObj a gradient step acts much like an implicit offset-free correction. On LeWM, $b^\ast$ helps where the shift hurts. Residual feedback gives no detectable gain there. Our experiments do not separate the other two readings. A second open question is why the remainder carries the gain under OpTTA's full objective (Sec.~6.4). Its planning-cost term acts on the candidates' predicted costs and may thus enlarge the remainder. Measuring how the remainder reorders the candidates would test this.

\paragraph{7.2 The shift as a lens}
With most of the gain running through one shared vector per replan, the shift also shows how the gain varies with search (Sec.~6.5) and with the base model (Sec.~6.1).

Residual feedback, which needs no parameter update, also gives a cheap reference for new update rules on PushObj. Adding and removing the shift separate whether the shift suffices for a gain from whether the gain needs it, which differ under OpTTA's full objective (Sec.~6.4). We suggest reporting the recovered fraction and the remainder gain at more than one search budget, alongside success rates.

\paragraph{7.3 Limitations}
We study CEM planning only. Our claims concern planners that score candidates by a latent goal-reaching cost. Planners with learned reward or value heads, as in JEPA-TTT, are outside them. The adaptation we study resets every episode, while updates that accumulate across episodes can lower held-out error. Each model has one released checkpoint, to which we add a further-trained ResNet base and four LeWM checkpoints of our own (Sec.~6.1). Tasks for an unseen shape often share their start configuration with tasks for seen shapes. On LeWM, OpTTA's protocol, which we follow, draws its tasks from the training demonstrations. Success requires both the pusher and the block to reach their goals. Counting the block alone cuts the PushObj gain by more than half (Sec.~5.4). On LeWM, where the frozen model already places the block in 96.2\% of episodes, the gain comes mostly from the pusher (post hoc).

\paragraph{7.4 Conclusion}
Test-time adaptation can raise planning success mostly by shifting the predictions of all candidates together, which is equivalent to re-aiming the planner at every replan. The shared shift makes this visible. We hope it helps future work read such gains.

{
    \small
    \bibliographystyle{ieeenat_fullname}
    \bibliography{main}
}

\clearpage
\includepdf[pages=-]{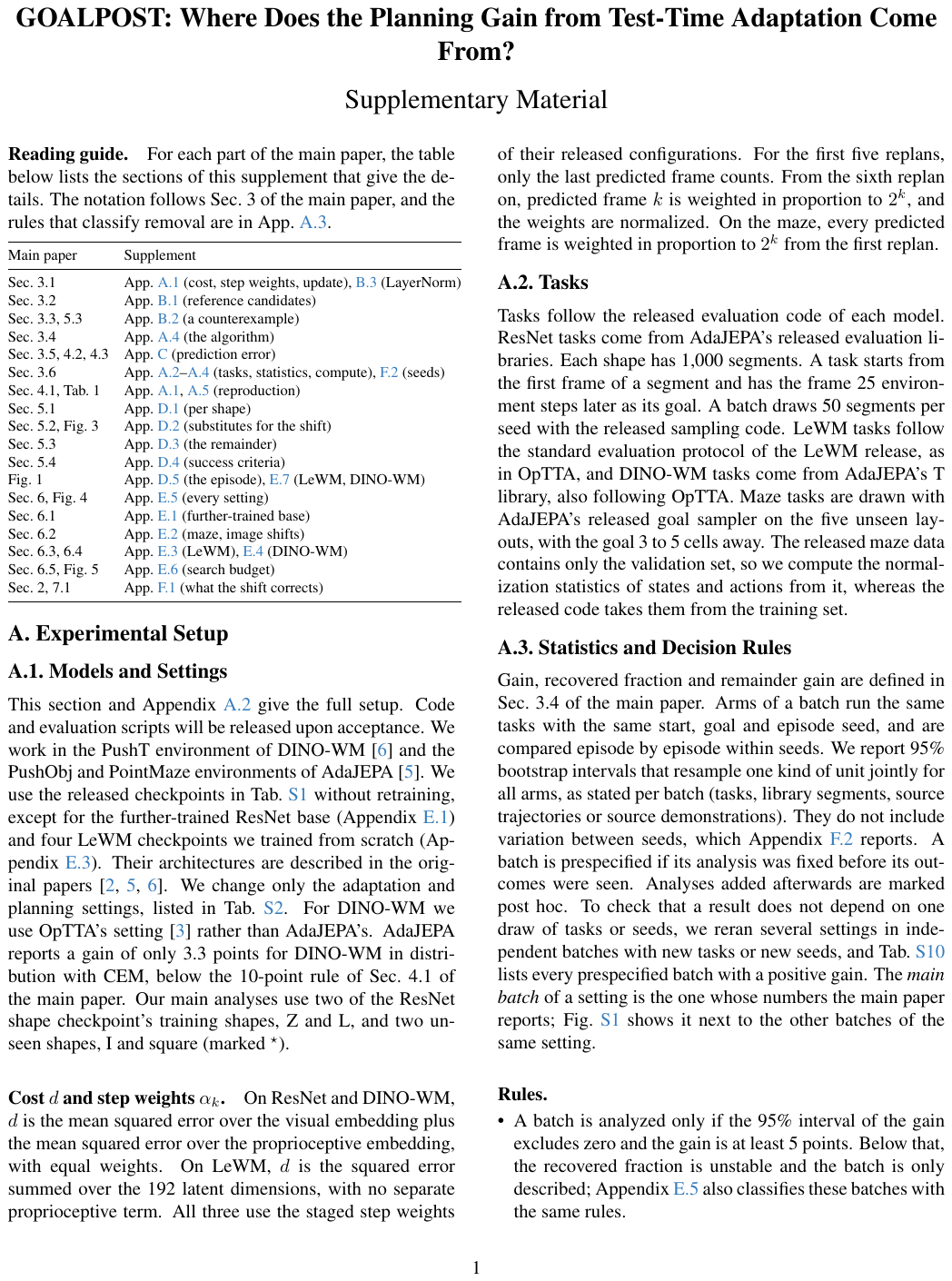}
\end{document}

%% file: preamble.tex
\usepackage{cuted} %

\usepackage{currfile} %

\usepackage{caption} %

\renewcommand{\paragraph}[1]{\vspace{.5em}\noindent\textbf{#1.}}

%% file: main.bbl
\begin{thebibliography}{37}
\providecommand{\natexlab}[1]{#1}
\providecommand{\url}[1]{\texttt{#1}}
\expandafter\ifx\csname urlstyle\endcsname\relax
  \providecommand{\doi}[1]{doi: #1}\else
  \providecommand{\doi}{doi: \begingroup \urlstyle{rm}\Url}\fi

\bibitem[Assran et~al.(2025)Assran, Bardes, Fan, Garrido, Howes, Komeili,
  Muckley, Rizvi, Roberts, Sinha, Zholus, Arnaud, Gejji, Martin, Hogan, Dugas,
  Bojanowski, Khalidov, Labatut, Massa, Szafraniec, Krishnakumar, Li, Ma,
  Chandar, Meier, LeCun, Rabbat, and Ballas]{assran2025vjepa2}
Mahmoud Assran, Adrien Bardes, David Fan, Quentin Garrido, Russell Howes,
  Mojtaba Komeili, Matthew Muckley, Ammar Rizvi, Claire Roberts, Koustuv Sinha,
  Artem Zholus, Sergio Arnaud, Abha Gejji, Ada Martin, Francois~Robert Hogan,
  Daniel Dugas, Piotr Bojanowski, Vasil Khalidov, Patrick Labatut, Francisco
  Massa, Marc Szafraniec, Kapil Krishnakumar, Yong Li, Xiaodong Ma, Sarath
  Chandar, Franziska Meier, Yann LeCun, Michael Rabbat, and Nicolas Ballas.
\newblock {V-JEPA} 2: Self-supervised video models enable understanding,
  prediction and planning.
\newblock \emph{arXiv preprint arXiv:2506.09985}, 2025.

\bibitem[Boney et~al.(2019)Boney, Di~Palo, Berglund, Ilin, Kannala, Rasmus, and
  Valpola]{boney2019regularizing}
Rinu Boney, Norman Di~Palo, Mathias Berglund, Alexander Ilin, Juho Kannala,
  Antti Rasmus, and Harri Valpola.
\newblock Regularizing trajectory optimization with denoising autoencoders.
\newblock In \emph{Advances in Neural Information Processing Systems
  (NeurIPS)}, 2019.

\bibitem[Chen et~al.(2026)Chen, Wang, and Li]{chen2026cqm}
Junlin Chen, Ruijie Wang, and Jianxin Li.
\newblock Counterfactual quotient models: Learning what actions change, not
  what the world does.
\newblock \emph{arXiv preprint arXiv:2608.22092}, 2026.

\bibitem[Cheung and He(2026)]{cheung2026onestep}
Joshua Cheung and Zhang He.
\newblock One-step prediction fidelity is a poor proxy for planning performance
  in {JEPA} world models.
\newblock \emph{arXiv preprint arXiv:2610.19375}, 2026.

\bibitem[Fu et~al.(2020)Fu, Kumar, Nachum, Tucker, and Levine]{fu2020d4rl}
Justin Fu, Aviral Kumar, Ofir Nachum, George Tucker, and Sergey Levine.
\newblock {D4RL}: Datasets for deep data-driven reinforcement learning.
\newblock \emph{arXiv preprint arXiv:2004.07219}, 2020.

\bibitem[Krause et~al.(2018)Krause, Kahembwe, Murray, and
  Renals]{krause2018dynamic}
Ben Krause, Emmanuel Kahembwe, Iain Murray, and Steve Renals.
\newblock Dynamic evaluation of neural sequence models.
\newblock In \emph{International Conference on Machine Learning (ICML)}, 2018.

\bibitem[Lambert et~al.(2020)Lambert, Amos, Yadan, and
  Calandra]{lambert2020objective}
Nathan Lambert, Brandon Amos, Omry Yadan, and Roberto Calandra.
\newblock Objective mismatch in model-based reinforcement learning.
\newblock In \emph{Learning for Dynamics and Control (L4DC)}, 2020.

\bibitem[LeCun(2022)]{lecun2022path}
Yann LeCun.
\newblock A path towards autonomous machine intelligence.
\newblock OpenReview, 2022.
\newblock Version 0.9.2, 2022-06-27.
  \url{https://openreview.net/forum?id=BZ5a1r-kVsf}.

\bibitem[Levy et~al.(2026)Levy, Westenbroek, Huang, Palafox, Yin, Omidshafiei,
  Kim, Gupta, and Fridovich-Keil]{levy2026simdist}
Jacob Levy, Tyler Westenbroek, Kevin Huang, Fernando Palafox, Patrick Yin,
  Shayegan Omidshafiei, Dong-Ki Kim, Abhishek Gupta, and David Fridovich-Keil.
\newblock Simulation distillation: Pretraining world models in simulation for
  rapid real-world adaptation.
\newblock In \emph{Robotics: Science and Systems (RSS)}, 2026.

\bibitem[Li et~al.(2026)Li, Fei, Zhou, and Hayashi]{li2026dueling}
Jiazhuo Li, Yiming Fei, Zhiruo Zhou, and Heikichi Hayashi.
\newblock Dueling world models: Advantage-style action channels for common-mode
  distractor rejection.
\newblock \emph{arXiv preprint arXiv:2608.06706}, 2026.

\bibitem[Maes et~al.(2026)Maes, Le~Lidec, Scieur, LeCun, and
  Balestriero]{maes2026lewm}
Lucas Maes, Quentin Le~Lidec, Damien Scieur, Yann LeCun, and Randall
  Balestriero.
\newblock {LeWorldModel}: Stable end-to-end joint-embedding predictive
  architecture from pixels.
\newblock \emph{arXiv preprint arXiv:2603.19312}, 2026.

\bibitem[Mete et~al.(2026)Mete, Kalathil, and Kumar]{mete2026optta}
Akshay Mete, Dileep Kalathil, and P.~R. Kumar.
\newblock Adapt for planning, not just prediction: Optimistic test-time world
  model adaptation.
\newblock \emph{arXiv preprint arXiv:2610.35636}, 2026.

\bibitem[Min et~al.(2026)Min, Li, Xu, and Du]{min2026lowerror}
Rui Min, Xianyao Li, Fang Xu, and Eric~Jing Du.
\newblock When low prediction error misleads planning: Diagnosing
  representation, dynamics, and decision failures in latent world models.
\newblock \emph{arXiv preprint arXiv:2610.05550}, 2026.

\bibitem[Muske and Badgwell(2002)]{muske2002offsetfree}
Kenneth~R. Muske and Thomas~A. Badgwell.
\newblock Disturbance modeling for offset-free linear model predictive control.
\newblock \emph{Journal of Process Control}, 12\penalty0 (5):\penalty0
  617--632, 2002.

\bibitem[Nado et~al.(2020)Nado, Padhy, Sculley, D'Amour, Lakshminarayanan, and
  Snoek]{nado2020evaluating}
Zachary Nado, Shreyas Padhy, D. Sculley, Alexander D'Amour, Balaji
  Lakshminarayanan, and Jasper Snoek.
\newblock Evaluating prediction-time batch normalization for robustness under
  covariate shift.
\newblock \emph{arXiv preprint arXiv:2006.10963}, 2020.

\bibitem[Nagabandi et~al.(2019{\natexlab{a}})Nagabandi, Clavera, Liu, Fearing,
  Abbeel, Levine, and Finn]{nagabandi2019grbal}
Anusha Nagabandi, Ignasi Clavera, Simin Liu, Ronald~S. Fearing, Pieter Abbeel,
  Sergey Levine, and Chelsea Finn.
\newblock Learning to adapt in dynamic, real-world environments through
  meta-reinforcement learning.
\newblock In \emph{International Conference on Learning Representations
  (ICLR)}, 2019{\natexlab{a}}.

\bibitem[Nagabandi et~al.(2019{\natexlab{b}})Nagabandi, Finn, and
  Levine]{nagabandi2019mole}
Anusha Nagabandi, Chelsea Finn, and Sergey Levine.
\newblock Deep online learning via meta-learning: Continual adaptation for
  model-based {RL}.
\newblock In \emph{International Conference on Learning Representations
  (ICLR)}, 2019{\natexlab{b}}.

\bibitem[Oquab et~al.(2024)Oquab, Darcet, Moutakanni, Vo, Szafraniec, Khalidov,
  Fernandez, Haziza, Massa, El-Nouby, Assran, Ballas, Galuba, Howes, Huang, Li,
  Misra, Rabbat, Sharma, Synnaeve, Xu, Jegou, Mairal, Labatut, Joulin, and
  Bojanowski]{oquab2024dinov2}
Maxime Oquab, Timoth{\'e}e Darcet, Th{\'e}o Moutakanni, Huy~V. Vo, Marc
  Szafraniec, Vasil Khalidov, Pierre Fernandez, Daniel Haziza, Francisco Massa,
  Alaaeldin El-Nouby, Mahmoud Assran, Nicolas Ballas, Wojciech Galuba, Russell
  Howes, Po-Yao Huang, Shang-Wen Li, Ishan Misra, Michael Rabbat, Vasu Sharma,
  Gabriel Synnaeve, Hu Xu, Herv{\'e} Jegou, Julien Mairal, Patrick Labatut,
  Armand Joulin, and Piotr Bojanowski.
\newblock {DINOv2}: Learning robust visual features without supervision.
\newblock \emph{Transactions on Machine Learning Research (TMLR)}, 2024.

\bibitem[Pannocchia and Rawlings(2003)]{pannocchia2003offsetfree}
Gabriele Pannocchia and James~B. Rawlings.
\newblock Disturbance models for offset-free model-predictive control.
\newblock \emph{AIChE Journal}, 49\penalty0 (2):\penalty0 426--437, 2003.

\bibitem[Qiu et~al.(2026)Qiu, Chen, Cao, Shi, Huo, and Gao]{qiu2026adwm}
Jiabin Qiu, Zixuan Chen, Hongye Cao, Jieqi Shi, Jing Huo, and Yang Gao.
\newblock {AD-WM}: Action-discriminative world models for counterfactual model
  predictive control.
\newblock \emph{arXiv preprint arXiv:2609.30264}, 2026.

\bibitem[Resani and Shi(2026)]{resani2026dojepa}
Hossein Resani and Javen~Qinfeng Shi.
\newblock {Do-JEPA}: From masking to intervention in latent world models.
\newblock \emph{arXiv preprint arXiv:2609.37378}, 2026.

\bibitem[Rubinstein and Kroese(2004)]{rubinstein2004cem}
Reuven~Y. Rubinstein and Dirk~P. Kroese.
\newblock \emph{The Cross-Entropy Method: A Unified Approach to Combinatorial
  Optimization, Monte-Carlo Simulation, and Machine Learning}.
\newblock Springer, 2004.

\bibitem[Schneider et~al.(2020)Schneider, Rusak, Eck, Bringmann, Brendel, and
  Bethge]{schneider2020improving}
Steffen Schneider, Evgenia Rusak, Luisa Eck, Oliver Bringmann, Wieland Brendel,
  and Matthias Bethge.
\newblock Improving robustness against common corruptions by covariate shift
  adaptation.
\newblock In \emph{Advances in Neural Information Processing Systems
  (NeurIPS)}, 2020.

\bibitem[Sobal et~al.(2025)Sobal, Zhang, Cho, Balestriero, Rudner, and
  LeCun]{sobal2025pldm}
Vlad Sobal, Wancong Zhang, Kyunghyun Cho, Randall Balestriero, Tim G.~J.
  Rudner, and Yann LeCun.
\newblock Learning from reward-free offline data: A case for planning with
  latent dynamics models.
\newblock In \emph{Advances in Neural Information Processing Systems
  (NeurIPS)}, 2025.

\bibitem[Song and Lee(2026)]{song2026dither}
Yong~Hwi Song and Keon~Myung Lee.
\newblock Planners exploit deterministic error: Dithered low-bit world models.
\newblock \emph{arXiv preprint arXiv:2611.11780}, 2026.

\bibitem[Soni et~al.(2026{\natexlab{a}})Soni, Sehgal, Dave, and
  Rueckert]{soni2026alra}
Krishnam Soni, Aditya Sehgal, Vedant Dave, and Elmar Rueckert.
\newblock {ALRA}: Action-latent residual adaptation for world models.
\newblock In \emph{NeurIPS Workshop on PTA: From Pretrained Representations to
  Acting Agents}, 2026{\natexlab{a}}.

\bibitem[Soni et~al.(2026{\natexlab{b}})Soni, Sehgal, Dave, and
  Rueckert]{soni2026sandwich}
Krishnam Soni, Aditya Sehgal, Vedant Dave, and Elmar Rueckert.
\newblock {Sandwich-Residuals}: Parameter-efficient test-time adaptation of
  world models.
\newblock \emph{arXiv preprint arXiv:2609.21740}, 2026{\natexlab{b}}.

\bibitem[Suliga et~al.(2026)Suliga, Vanjani, Nilavadi, Burgard, and
  Lioutikov]{suliga2026residual}
Jakub Suliga, Pankhuri Vanjani, Nisarga Nilavadi, Wolfram Burgard, and Rudolf
  Lioutikov.
\newblock Few-shot residual adaptation of pretrained latent world models.
\newblock \emph{arXiv preprint arXiv:2611.13592}, 2026.

\bibitem[Thrun et~al.(1990)Thrun, M{\"o}ller, and Linden]{thrun1990adaptive}
Sebastian~B. Thrun, Knut M{\"o}ller, and Alexander Linden.
\newblock Planning with an adaptive world model.
\newblock In \emph{Advances in Neural Information Processing Systems (NIPS)},
  pages 450--456, 1990.

\bibitem[Tian et~al.(2023)Tian, Finn, and Wu]{tian2023vp2}
Stephen Tian, Chelsea Finn, and Jiajun Wu.
\newblock A control-centric benchmark for video prediction.
\newblock In \emph{International Conference on Learning Representations
  (ICLR)}, 2023.

\bibitem[Wang et~al.(2021)Wang, Shelhamer, Liu, Olshausen, and
  Darrell]{wang2021tent}
Dequan Wang, Evan Shelhamer, Shaoteng Liu, Bruno Olshausen, and Trevor Darrell.
\newblock Tent: Fully test-time adaptation by entropy minimization.
\newblock In \emph{International Conference on Learning Representations
  (ICLR)}, 2021.

\bibitem[Wang et~al.(2026{\natexlab{a}})Wang, Bounou, LeCun, and
  Ren]{wang2026adajepa}
Ying Wang, Oumayma Bounou, Yann LeCun, and Mengye Ren.
\newblock {AdaJEPA}: An adaptive latent world model.
\newblock \emph{arXiv preprint arXiv:2606.32026}, 2026{\natexlab{a}}.

\bibitem[Wang et~al.(2026{\natexlab{b}})Wang, Bounou, Zhou, Balestriero,
  Rudner, LeCun, and Ren]{wang2026straightening}
Ying Wang, Oumayma Bounou, Gaoyue Zhou, Randall Balestriero, Tim G.~J. Rudner,
  Yann LeCun, and Mengye Ren.
\newblock Temporal straightening for latent planning.
\newblock In \emph{International Conference on Machine Learning (ICML)},
  2026{\natexlab{b}}.

\bibitem[Zhang et~al.(2026{\natexlab{a}})Zhang, Jiang, Dai, Lu, Uzunoglu,
  Zhang, Wei, Wang, Patel, Liang, Khashabi, Peng, Chellappa, Shu, Yuille, Du,
  and Chen]{zhang2026worldinworld}
Jiahan Zhang, Muqing Jiang, Nanru Dai, Taiming Lu, Arda Uzunoglu, Shunchi
  Zhang, Yana Wei, Jiahao Wang, Vishal~M. Patel, Paul~Pu Liang, Daniel
  Khashabi, Cheng Peng, Rama Chellappa, Tianmin Shu, Alan Yuille, Yilun Du, and
  Jieneng Chen.
\newblock {World-in-World}: World models in a closed-loop world.
\newblock In \emph{International Conference on Learning Representations
  (ICLR)}, 2026{\natexlab{a}}.

\bibitem[Zhang et~al.(2026{\natexlab{b}})Zhang, Shi, Peng, and
  Xie]{zhang2026efe}
Tong Zhang, Zhiyuan Shi, Yun Peng, and Tao Xie.
\newblock Knowing when to stop thinking: A dissection of expected free energy
  as a test-time compute controller for latent world-model planning.
\newblock \emph{arXiv preprint arXiv:2610.28011}, 2026{\natexlab{b}}.

\bibitem[Zhang et~al.(2026{\natexlab{c}})Zhang, Ye, Agarwal, Mahjoub,
  Moradi~Pari, Khashabi, Shu, and Tadiparthi]{zhang2026jepattt}
Zheyuan Zhang, Suyu Ye, Nakul Agarwal, Hossein~Nourkhiz Mahjoub, Ehsan
  Moradi~Pari, Daniel Khashabi, Tianmin Shu, and Vaishnav Tadiparthi.
\newblock {JEPA-TTT}: Persistent test-time training of latent world models for
  planning under dynamics shifts.
\newblock \emph{arXiv preprint arXiv:2610.00722}, 2026{\natexlab{c}}.

\bibitem[Zhou et~al.(2025)Zhou, Pan, LeCun, and Pinto]{zhou2025dinowm}
Gaoyue Zhou, Hengkai Pan, Yann LeCun, and Lerrel Pinto.
\newblock {DINO-WM}: World models on pre-trained visual features enable
  zero-shot planning.
\newblock In \emph{International Conference on Machine Learning (ICML)}, 2025.

\end{thebibliography}
